\documentclass{article}

\usepackage[final]{neurips_2022}

\usepackage[utf8]{inputenc} % allow utf-8 input
\usepackage[T1]{fontenc}    % use 8-bit T1 fonts
\usepackage{hyperref}       % hyperlinks
\usepackage{url}            % simple URL typesetting
\usepackage{booktabs}       % professional-quality tables
\usepackage{amsfonts}       % blackboard math symbols
\usepackage{nicefrac}       % compact symbols for 1/2, etc.
\usepackage{microtype}      % microtypography
\usepackage{xcolor}         % colors
\usepackage{amsmath,amssymb,amsthm}
\usepackage{algorithmic}
\usepackage{graphicx}
\usepackage{textcomp}
\usepackage{multirow,array,booktabs}
\usepackage{placeins}
\usepackage{url}

\newtheorem{thm}{Theorem}
\theoremstyle{definition}
\newtheorem{definition}{Definition}

\title{Time Series Causal Link Estimation under Hidden Confounding using Knockoff Interventions}

\author{%
  Violeta Teodora Trifunov \\
  Department for Mathematics and Computer Science, Computer Vision Group\\
  Friedrich Schiller University Jena\\
  Jena, Germany\\
  \texttt{violetateodora.trifunov@uni-jena.de} \\
   \AND
   Maha Shadaydeh \\
   Department for Mathematics and Computer Science, Computer Vision Group\\
   Friedrich Schiller University Jena\\
   Jena, Germany\\
   \AND
   Joachim Denzler \\
   Department for Mathematics and Computer Science, Computer Vision Group\\
   Friedrich Schiller University Jena\\
   Jena, Germany \\
}

\begin{document}

\maketitle

\begin{abstract}
  Latent variables often mask cause-effect relationships in observational data which provokes spurious links that may be misinterpreted as causal. This problem sparks great interest in the fields such as climate science and economics. We propose to estimate confounded causal links of time series using Sequential Causal Effect Variational Autoencoder (SCEVAE) while applying Knockoff interventions. Knockoff variables have the same distribution as the originals and preserve the correlation to other variables. This allows for counterfactuals that are more faithful to the observational distribution. We show the advantage of Knockoff interventions by applying SCEVAE to synthetic datasets with both linear and nonlinear causal links. Moreover, we apply SCEVAE with Knockoffs to real aerosol-cloud-climate observational time series data. We compare our results on synthetic data to those of a time series deconfounding method both with and without estimated confounders. We show that our method outperforms this benchmark by comparing both methods to the ground truth. For the real data analysis, we rely on expert knowledge of causal links and demonstrate how using suitable proxy variables improves the causal link estimation in the presence of hidden confounders.
\end{abstract}

\section{Introduction}\label{intro}

Causal link estimation and inference for dynamical systems are important tasks in many fields such as finance and climate science \citep{Jakob19}, \citep{Peters}. We propose a novel method for estimating both linear and nonlinear causal effects in time series under hidden confounding while performing Knockoff \citep{knockoffs} interventions, inspired by the Causal Effect Variational Autoencoder (CEVAE) \citep{CEVAE}. The proposed approach, as will be explained further on, allows for less biased causality analysis of non-stationary sequential data and counterfactuals that are more faithful to the distribution of observational data. A hidden confounder is a variable that influences both the cause and the effect, and its presence can lead to false relationships which may be misconceived as causal. 

\begin{figure}[t]
	\centering
	\includegraphics[width=0.3\textwidth]{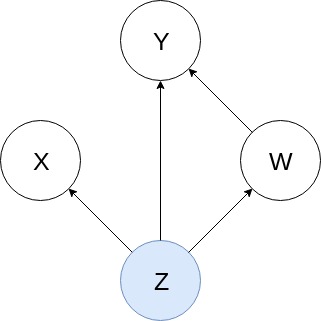}
	\caption{Causal DAG with observed variables $\{ W, X, Y\}$  and latent variable $\{ Z\}$.}
	\label{fig:CEVAE_model}
\end{figure}

CEVAE is a deep learning framework which was first applied to analyzing effects of a binary treatment on the patients' health outcomes. The causal Directed Acyclic Graph (DAG) representing hidden confounding with one proxy variable, as used by CEVAE framework, is depicted in Fig. \ref{fig:CEVAE_model}. $Y$ denotes the effect, $W$ the cause, $Z$ denotes the hidden confounder and $X$ a proxy variable providing noisy views on $Z$. We note that the proxy can be multivariate, and both categorical and continuous. Although the DAG in Fig. \ref{fig:CEVAE_model} might appear restrictive, in many real-world applications, such as climate science for instance, there are unobserved confounders present, with multiple different proxy variables to describe them \citep{CI18_hidden_conf}.
We assume that the causal link between the confounded variables is approximately directly identifiable through the proxy to the hidden confounder. We provide more details on the identifiability theory in Section \ref{sec:identifiability}.

Knockoff filters were first introduced by \cite{knockoffs} for controlling the variable selection procedure and false discovery rate of potential drivers of the response variable. Their method produces Knockoff variables which imitate the correlation structure of the existing variables and do not require any new data. At first they were limited to a linear Gaussian response variable, but this was later extended for arbitrary and unobserved data distributions in the work by \cite{deep_knockoffs}, as they introduced a framework called Deep Knockoffs based on deep generative models.

Due to its highly practicable properties for counterfactual analysis \citep{Pearl}, as proposed by \cite{Wasim_ICMLW}, we generate a Knockoff \citep{knockoffs} of the cause variable $W$ as a way of intervention to estimate the causal link intensity using our approach. In addition to preserving the distribution of $W$, this also preserves the inter-variable correlations to other observed variables, but is not correlated to $W$. To highlight the advantage of Knockoff intervention, we compare it to using Gaussian noise as intervention variable $\widehat{W}$. It is independent of the hidden confounder $Z$ and therefore removes the link between $W$ and $Z$ as required by $do$-calculus \citep{Pearl}. However, Gaussian noise intervention does not preserve the distribution of the original cause variable $W$, which is learned during the training of the neural networks within our method and expected for inference. This leads to suboptimal performance in contrast to Knockoff interventions.

Apart from the nonlinear causal links, the problem of causal link estimation under hidden confounding is particularly challenging for time series because the data is non-stationary, and the causal link between confounded variables might be changing over time, thus introducing delayed causal effects.

To address these issues, we engineered our approach to transcend the CEVAE methodology and become applicable to complicated non-stationary sequential data with the help of Long Short-Term Memory (LSTM) \citep{LSTM} recurrent neural networks. Moreover, we allow for the causal effect estimation to be done on a single time series example of each observed variable. This means that one does not need to have multiple realizations of the variables on their disposal which is an advantage in many fields such as finance and climate science where each variable is observed only once per time step.

The paper is structured as follows. In Section \ref{sec:rel_work}, we outline related work on treatment effect estimation and provide an overview of causality analysis under hidden confounding. Moreover, we discuss other associated work that utilizes Knockoffs. In Section \ref{sec:method}, we introduce the CEVAE framework, define and discuss causal identifiability, the assumptions under which using SCEVAE is justified, as well as a detailed description of our method, and introduce Knockoffs themselves in more detail. Section \ref{sec:data} describes the data we use, and Section \ref{sec:experiments} provides a detailed experimental setup along with our method's results. Section \ref{sec:conclusion} summarizes our approach and concludes the paper.

To the best of our knowledge, we are the first to propose a causal effect estimation method under hidden confounding for sequential data based on CEVAE and utilize it with Knockoff interventions.

\section{Related Work}\label{sec:rel_work}

Treatment effect estimation in the static setting is a very active field of research \citep{Bayes_ITE}, \citep{SVD_ITE_hidd_conf}, \citep{SyncTwin}. Under the assumption of no hidden confounding, methods such as VCNet \citep{VCnet} and its extensions by state-of-the-art transformer networks \citep{transformers} were recently applied to the task of continuous treatment estimation by \cite{TransTEE} and \cite{causal_transformer}.

In a more realistic scenario when there are unobserved confounders present, Causal Effect Variational Autoencoder (CEVAE) \citep{CEVAE} was one of the first such deep learning methods. It relies on the existence of one or more proxies to directly estimate the latent confounder. Nevertheless, in contrast to above-mentioned transformer-based treatment estimation methods, CEVAE focuses on binary treatment.
Research endeavors such as those by \cite{CEVAE_critic}, \cite{CEVAE_uniform_t}, and \cite{TVT_GCPR19} either extend CEVAE to a uniform or continuous treatment, or analyse this deep latent variable model's capabilities. However, until now, to the best of our knowledge, CEVAE has never been applied to time series.

Taking a different approach towards treatment effect estimation, \cite{Bica} and \cite{SeqDeconf} propose deconfounding methods for sequential data using neural networks. Specifically, in the work by \cite{Bica}, the authors develop Time Series Deconfounder (TSdeconf), a method built upon RNNs with multitask output to produce a factor model over time and estimate latent variables. These latent variable estimates are then used for causal inference as proxies. The limitation of this approach, in contrast to ours, is that it requires many patients i.e. samples and is not suitable for processing long time series (e.g. $N \geq 1000$).
Another shortcoming of TSdeconf is that it cannot be applied to observational data with one or more treatments assigned at each time step. This was addressed by \cite{SeqDeconf}, as the authors introduced a similar time series deconfounding method called Sequential Deconfounder, this time based on a Gaussian process latent variable model.

These methods are useful when there are no proxies available. In contrast to SCEVAE, this may introduce more approximation error and requires multiple realizations of each variable i.e. many independent samples for training. Although SCEVAE relies on access to proxies, it is not a limitation in many fields such as environmental and climate science, where many observed variables can be used to describe a latent one \citep{CI18_hidden_conf}.

In the recent years, due to their distribution- and correlation-preserving properties, Knockoffs \citep{knockoffs}, \citep{deep_knockoffs} were innovatively applied for counterfactual analysis of images \citep{Oana}, and time series \citep{Wasim}, \citep{Wasim_ICMLW}. \cite{Wasim} propose to use deep learning and counterfactual Knockoff variables to aid causal discovery of multivariate nonlinear time series with lower false discovery rate than state-of-the-art methods.

Similarly to SCEVAE, work by \cite{GrangerGRU} relies on RNNs to estimate the causal link of the variables under influence of a hidden confounder. However, it does not employ do-calculus \citep{Pearl}, but rather Granger causality \citep{GC} to determine the presence of a non-spurious causal link. Due to this choice of causality analysis tools, it cannot detect instantaneous causal links. Moreover, since we use knockoff interventions the cause variable's distribution does not change, and its correlation to other observation variables is preserved. This type of intervention allows for more accurate causal link estimates in comparison to standard normal interventions, as well as lower counterfactual prediction error as will be discussed in more detail in Section \ref{sec:results}.

\section{Methodology}\label{sec:method}

\subsection{Causal effect variational autoencoder}\label{sec:CEVAE}

To better understand SCEVAE architecture, we first introduce Causal effect variational autoencoder (CEVAE) \citep{CEVAE}. It is a deep learning method based on a VAE \citep{VAE} and a TARnet \citep{TARnet}. Its underlying probabilistic graphical model is shown in Fig. \ref{fig:CEVAE_model}.
CEVAE methodology assumes all variables to be non-sequential. $W$ denotes binary treatment, $Y$ an outcome of this treatment, while the latent confounder $Z$, and its proxy $X$ denote the socio-economic status and income of each patient, respectively. The central aim of treatment effect estimation is recovering the Individual Treatment Effect (ITE) and the Average Treatment Effect (ATE) defined in (\ref{eq:ITE}) and (\ref{eq:ATE}), respectively:
%For a discrete proxy, the difference of the intensity of a causal link between $W$ and $Y$ before and after intervention in $W$, in equation (\ref{eq:ITE}) denoted by $w^{0}$ and $w^{1}$ respectively, is measured by $ITE$, as well as $ATE$:
\begin{align}\label{eq:ITE}
	\begin{split}
		\text{ITE}(x) := &\mathbb{E}_{Y}(Y|X=x,do(W=w^{1})) - \mathbb{E}_{Y}(Y|X=x,do(W=w^{0}))
	\end{split}
\end{align}
\begin{equation}\label{eq:ATE}
	\text{ATE} :=  \ \mathbb{E}_{Y}(ITE(x))
\end{equation}
These metrics are defined for each value $x$ of variable $X$, and by $w^{1}$ we denote applied treatment, while values of $W$ when no treatment is applied are denoted by $w^{0}$. In the CEVAE framework, $w^{1} = 1$, and $w^{0} = 0$.
ATE is easily calculated once we obtain the ITE, and for that we need to recover the joint distribution $p(Z,X,W,Y)$, as stated in Theorem~1 by \cite{CEVAE}.
\begin{thm}
If CEVAE recovers $p(Z,X,W,Y)$, then we can recover the ITE under the causal model in Fig. \ref{fig:CEVAE_model}.
\end{thm}
Distribution $p(Z,X,W,Y)$ is obtained via CEVAE's model network by approximating the true posterior over $Z$ conditioned on $X$, $W$ and $Y$, whereas the prior $p(Z)$ is modeled by the standard normal distribution. All estimated probability distributions are parameterized by MLPs. TARnet is used to infer the estimate of the posterior by branching for each of the two treatment groups in $W$.
% Further, once all necessary distributions have been modeled, one can then construct a single objective for the inference and model networks, i.e. the \textit{variational lower bound}:
% \begin{align}\label{eq:var_lower_bound}
% 	\begin{split}
% 		\mathcal{L} =
% 		\sum_{i=1}^{N} & \mathbb{E}_{q(z_{i}|x_{i},w_{i},y_{i})}(\log p(z_{i}) - \log q(z_{i}|x_{i},w_{i},y_{i}) + \log p(x_{i},w_{i}|z_{i}) + \log p(y_{i}|w_{i},z_{i}))
% 	\end{split}\raisetag{13pt}
% \end{align}
% \noindent of the causal graphical model from Fig. \ref{fig:CEVAE_model}. By $x_{i}$ we denote an input data point, by $w_{i}$ the treatment assignment, by $y_{i}$ the outcome of the specific treatment, by $z_{i}$ the latent confounder, and by $q(\cdot)$ we denote estimation of the probability distribution with the same arguments.

% Finally, since it is necessary to know the intervention assignment $W$ together with its outcome $Y$ before inferring the posterior distribution over $Z$, two auxiliary distributions are introduced, helping to predict $w_{i}$ and $y_{i}$ for new samples, so the variational lower bound becomes
% \begin{align}\label{eq:F_lower_bound}
% 	\begin{split}
% 		\mathcal{F}_{CEVAE}= \mathcal{L} + \sum_{i=1}^{N}(&\log q(w_{i}=w_{i}^{*}|x_{i}^{*}) + \log q(y_{i}=y_{i}^{*}|x_{i}^{*},w_{i}^{*})),
% 	\end{split}
% \end{align}

% \noindent where $x_{i}^{*}$, $w_{i}^{*}$, $y_{i}^{*}$ are the observed values for the input, intervention, and outcome variables in the training set.

\subsection{Causal identifiability}\label{sec:identifiability}

When estimating causal link intensity or performing causal discovery from observational data, one needs to establish if the underlying model is identifiable. If that is not the case, a set of assumptions under which the identifiability holds must be imposed.
% In the context of CEVAE, the main identifiability assumption is that the joint distribution $p(Z, X, W, Y)$ can be approximately obtained from the observations $(X, W, Y)$ alone \cite{CEVAE}. The authors further state multiple occasions when this is possible, such as when
% \begin{enumerate}
%     \item \label{ident_assumption} $X$ encompasses three independent realizations of $Z$,
%     \item  $Z$ is categorical and $X$ is a Gaussian mixture model with components regulated by $Z$, or
%     \item $Z$ consists of binary variables and $X$ is comprised of noisy-OR functions of $Z$.
% \end{enumerate}
% Since CEVAE was introduced, there have been advances in identifiability theory.
We introduce identifiability following recent work by \cite{identifiability} and discuss this issue for deep latent variable models such as VAE.
\begin{definition}\label{def:identifiability}
    Let $\sim$ be an equivalence relation on the set of model parameters $\Theta$. We say that a deep latent variable model $p_{\theta}(Y, Z) = p_{\theta}(Y|Z) p_{\theta}(Z)$, for observed variable $Y \in \mathbb{R}^{d}$ and latent random vector $Z \in \mathbb{R}^{n}$ is \textit{identifiable} up to $\sim$ if
    \begin{equation}
        p_{\theta}(Y) = p_{\hat{\theta}}(Y) \Rightarrow \theta \sim \hat{\theta}
    \end{equation}
    for $\theta, \hat{\theta} \in \Theta$. The elements of the quotient space $\Theta /_\sim$ are called the \textit{identifiability classes}.
\end{definition}
The graphical model in Fig. \ref{fig:CEVAE_model} is in general not identifiable. A starting point would be to obtain model parameters or estimate latent variables $\textbf{Z}^{*}$ up to trivial transformations $T_{i}$ for $i \in \{ 1, \dots, N\}$ in the form of sufficient statistics, and invertible matrix $A$ \citep{identifiability}.
% $\textbf{Z}^{*}$

To do that, we will first introduce sufficient statistics $T_{i}$ with respect to the deep latent variable model from Definition \ref{def:identifiability} and illustrate the procedure on a one-dimensional latent variable case.

A statistic is sufficient for a family of probability distributions if the sample from which it is obtained provides no additional information than the statistic, as to which of those probability distributions the data was sampled from \citep{suf_stats}.

Let $T_{i}=(T_{i,1}, \dots, T_{i,k})$, $i \in \{ 1, \dots, N\}$, $k \in \mathbb{N}$ be sufficient statistics with respect to the deep latent variable model associated with a standard normal family of parameters $\lambda_{i}(X) =(\lambda_{i,1}(X), \dots, \lambda_{i,k}(X))$, given a conditioning variable $X$. %The conditional probability distribution of the data, given its statistic, is independent of the parameter $\lambda_{i}(U)$.
Importantly, the conditioning variable $X$ is a proxy of the hidden confounder in our causal setup and functions $\lambda_{i}$ are parameterized by LSTMs.

To illustrate, as per \cite{identifiability}, let $k=1$, set $T_{i}:=T_{i,1}$, and let $A$ be an invertible matrix. We can then recover $\textbf{Z}$ related to the original $\textbf{Z}^{*}$ as follows:
\begin{equation}\label{def:suf_stats}
    (T_{1}^{*}(Z_{1}^{*}), \dots, T_{n}^{*}(Z_{n}^{*})) = A(T_{1}(Z_{1}), \dots, T_{1}(Z_{1})).
\end{equation}
\noindent This means we can estimate the original latent variables up to point-wise transformations $T_{i}^{*}, T_{i}$. 
In certain cases of non-sequential data, having $A$ as a permutation matrix reduces the problem of indeterminacy of $\textbf{Z}^{*}$ to finding the point-wise transformations of $\text{Z}$. This is due to Eq. (\ref{def:suf_stats}) then becoming $T_{i}^{*}(Z_{i}^{*}) = T_{i'}(Z_{i'})$ for a permuted index $i'$.

% We extend the identifiability theory of VAEs from \cite{identifiability} to CEVAE setup.
% Assumption 1: Prior $p_{\theta}(Z | X)$ is assumed to be conditionally factorial
Our deep latent variable model represented through parameters $\theta=(f, T_{i}, \lambda_{i})$, as per \cite{identifiability}, is
\begin{equation*}\label{eq:CEVAE_latent_model}
    p_{\theta}(Y, W, Z | X) = p_{f}(Y, W | Z) p_{T_{i}, \lambda_{i}}(Z | X),    
\end{equation*}
for a function $f$ parameterized by an LSTM and sufficient statistic $T_{i}$ with parameters $\lambda_{i}$, $i \in \{ 1, \dots, N\}$.
Some of the identifiability assumptions for a VAE as described by \cite{identifiability} are that function $f$ is injective, and that sufficient statistic $T_{i}$ is differentiable almost everywhere. Since our model is in principle a VAE, we rely on these assumptions and generate the synthetic data accordingly. We note, however, that in a dynamical-system setting this might not always suffice for the causal model to be identifiable and further theoretical work in this direction is required.

\subsection{Sequential CEVAE}\label{sec:SCEVAE}

\begin{figure*}[t]
	\centering
	\begin{minipage}{.5\textwidth}
	\centering
	\includegraphics[width=0.9\linewidth]{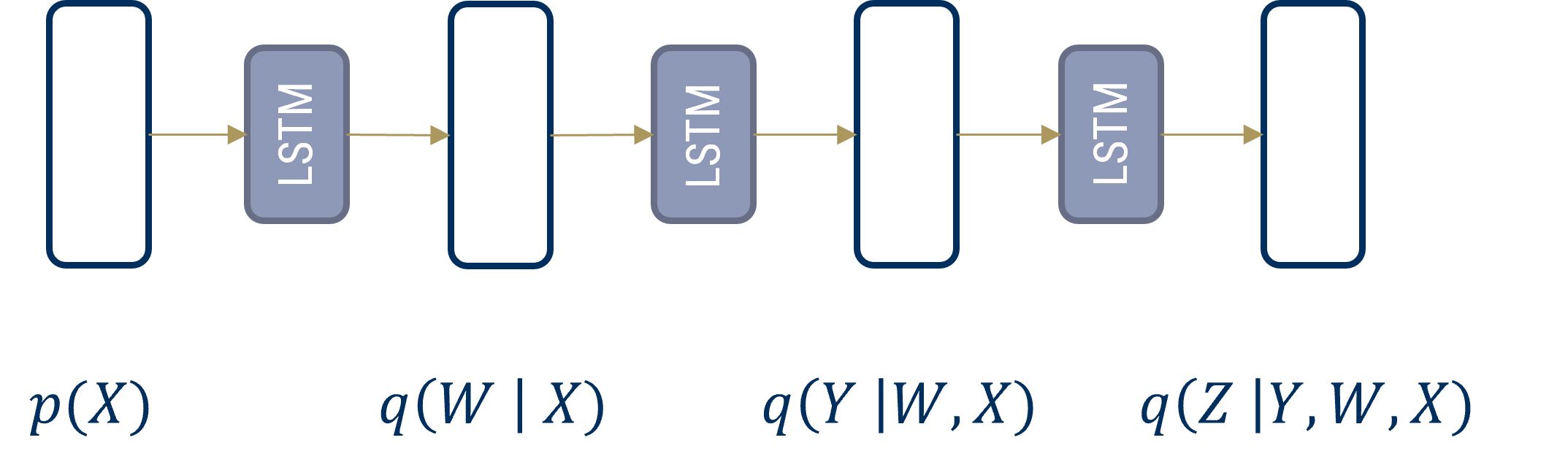}
%	\caption{SCEVAE encoder.}
	\end{minipage}%
	 \begin{minipage}{0.5\textwidth}
		\centering
		\includegraphics[width=0.9\linewidth]{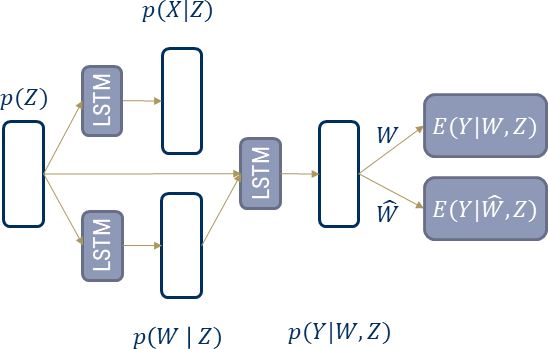}
%		\caption{SCEVAE decoder.}
	\end{minipage}
	\caption{SCEVAE architecture. The encoder is shown on the left, and the decoder with branching to estimate the factual and counterfactual outcomes on the right.}
	\label{fig:SCEVAE}
\end{figure*}

We propose a novel method for time series causal link estimation under hidden confounding. It is predicated on CEVAE, but differs significantly in the architecture and relies on LSTMs to model the involved probability distributions. Namely, we do not use a TARnet to estimate $p(Y | X, w^{0})$ and $p(Y | X, w^{1})$, specialized for binary treatment, but rather model $p(Y | X, W)$ sequentially, and introduce branching in the decoder as depicted on the right of Fig. \ref{fig:SCEVAE}. This branching allows us to compare factual and counterfactual causal effects, in order to estimate the cause-effect intensity after intervention on the non-binary sequential cause variable.
% Moreover, in Section \ref{sec:experiments} we show how our method can be used for causal discovery in real environmental data by varying cause and effect variables and using the rest of the available variables as proxies.

We will now introduce the notation for sequential data and explain the proposed method in full detail. Let $X = \{ x_{t} \}_{t=1}^{N}$, $W = \{ w_{t} \}_{t=1}^{N}$, and $Y = \{ y_{t} \}_{t=1}^{N}$ be the time series of proxy, cause variable, and effect, for $N \in \mathbb{N}$, respectively. The hidden confounder $Z = \{ z_{t} \}_{t=1}^{N}$ is also time series for each $z_{t} \in \mathbb{R}^{d}$, where $d$ is the dimension of $Z$ in the latent space at each time step $t$. The conditional distributions of these variables used by our method's framework are depicted in Fig. \ref{fig:SCEVAE}. The LSTMs are denoted by $f_{i}$, and each is parameterized by its own parameters $\phi_{i}$, for $i \in \{1, 2, 3, 4\}$.
For each time step $t$, we model $x_{t}$, $w_{t}$, and the prior of $z_{t}$ as follows:
\begin{align}
	p(z_{t}) &= \mathcal{N}(0,1)\\
	p(w_{t} | z_{t}) &= \mathcal{N}(\mu_{w_{t}}, \sigma_{w_{t}}^{2}), \ \ \ \mu_{w_{t}}, \sigma_{w_{t}}^{2} = f_{1}(z_{t})\\
	p(x_{t} | z_{t}) &= \mathcal{N}(\mu_{x_{t}}, \sigma_{x_{t}}^{2}), \ \ \ \mu_{x_{t}}, \sigma_{x_{t}}^{2} = f_{2}(z_{t}).
\end{align}
\noindent This means that each variable is true time series, having each individual time step modeled by a Gaussian distribution. The LSTMs are not bidirectional, hence they do not have access to future values of the input variable. This will be tackled in the future work.

When proxy $X$ contains binary components, we model them as $p(x_{t} | z_{t}) = \text{Bern}(\pi = \sigma(v(z_{t})))$, for the sigmoid function $\sigma$, and $v: \mathbb{R} \xrightarrow{} \mathbb{R}$, a real-valued function parameterized by an LSTM.

The mean and variance of the effect $y_{t}$ at each time step $t$ are also parameterized by an LSTM:
\begin{equation}\label{eq:effect_SCEVAE}
	p(y_{t} | w_{t}, z_{t}) = \mathcal{N}(\mu_{y_{t}}, \sigma^{2}_{y_{t}}), \ \ \ \mu_{y_{t}}, \sigma^{2}_{y_{t}} = f_{3}(w_{t}, z_{t})
\end{equation}
We indicate that in contrast to CEVAE, we do not fix variance in Eq. (\ref{eq:effect_SCEVAE}) but rather learn it from observational data.
% Since $W$ is continuous and a time series, we model this distribution at each time step, thus avoiding branching our method's architecture for different values of $W$, in contrast to CEVAE framework.
Our approach is less restrictive in the sense of its compatibility with continuous, sequential cause variable, and can therefore be applied to a wider variety of real-world problems. We use a Knockoff \citep{knockoffs} of $W$ as intervention since it preserves the original distribution of the data. This is of particular importance for creating counterfactuals because the trained neural network anticipates test and training data stemming from the same distribution. For comparison, we also apply Gaussian noise as intervention on $W$ which removes the link to the hidden confounder but does not preserve the original distribution of $W$. One limitation of our approach is that we assume that, within the processed window, data is stationary and sufficient to produce good Knockoffs.

According to the DAG in Fig. \ref{fig:CEVAE_model}, the posterior distribution of $Z$ depends on $X$, $Y$, and $W$. We thus approximate it by:
\begin{equation}
	q(z_{t} | x_{t}, y_{t}, w_{t}) = \mathcal{N}(\mu_{z_{t}}, \sigma_{z_{t}}^{2}), \ \mu_{z_{t}}, \sigma_{z_{t}}^{2} = f_{4}(x_{t}, y_{t}, w_{t}).
\end{equation}
To estimate the parameters of the auxiliary distribution of $W$ and $Y$ in Eq. (\ref{eq:F_lower_bound}), we use the following:
\begin{align}
	q(w_{t} | x_{t}) &= \mathcal{N}(\mu^{*}_{w_{t}}, {\sigma^{*}_{w_{t}}}^{2}), \ \mu^{*}_{w_{t}}, {\sigma^{*}}_{w_{t}}^{2} = f_{5}(x_{t})\\
	q(y_{t} | x_{t}, w_{t}) &= \mathcal{N}(\mu^{*}_{y_{t}}, {\sigma^{*}}_{y_{t}}^{2}), \mu^{*}_{y_{t}}, {\sigma^{*}}^{2}_{y_{t}} = f_{6}(x_{t}, w_{t})
\end{align}
\noindent To insure that variances are positive, we apply a softplus activation function $\textit{softplus}(u) = \ln{(1  + e^{u})}$, for $u \in \mathbb{R}$, to the output of a given LSTM cell used to parameterize the variance.

% By $x_{i}$ we denote an input data point, by $w_{i}$ the treatment assignment, by $y_{i}$ the outcome of the specific treatment, by $z_{i}$ the latent confounder, and by $q(\cdot)$ we denote estimation of the probability distribution with the same arguments.

We weigh the regularization loss by $\lambda \in \mathbb{R}$ in order to obtain more stable effect predictions, so the variational lower bound involving all modeled distributions is:
\begin{align}\label{eq:SCEVAE_var_lower_bound}
	\begin{split}
		\mathcal{\hat{L}} =
		\sum_{i=1}^{N} & \mathbb{E}_{q(z_{t}|x_{t},w_{t},y_{t})}(\lambda (\log p(z_{t}) - \log q(z_{t}|x_{t},w_{t},y_{t})) + \log p(x_{t},w_{t}|z_{t}) + \log p(y_{t}|w_{t},z_{t})).
	\end{split}\raisetag{13pt}
\end{align}
Similarly to CEVAE, since it is necessary to know the intervention assignment $W$ together with its outcome $Y$ before inferring the posterior distribution over $Z$, two auxiliary distributions are introduced, helping to predict $w_{t}$ and $y_{t}$ for new samples.
The variational lower bound used as an objective for both the inference and the model networks of our method is then:
\begin{align}\label{eq:F_lower_bound}
	\begin{split}
		\mathcal{F}_{\text{SCEVAE}}= \mathcal{\hat{L}} + \sum_{i=1}^{N}(&\log q(w_{t}=w_{t}^{*}|x_{t}^{*})+\log q(y_{t}=y_{t}^{*}|x_{t}^{*},w_{t}^{*})),
	\end{split}
\end{align}
for $x_{t}^{*}$, $w_{t}^{*}$, $y_{t}^{*}$ being the observed values for the input, intervention, and outcome variables in the training set.

Our method's encoder and decoder are depicted in Fig. \ref{fig:SCEVAE}. We call this novel approach Sequential Causal Effect Variational Autoencoder (SCEVAE, pronounced /see-VAE/).

\subsection{Knockoffs}\label{sec:knockoffs}

The idea of Knockoffs was introduced by \cite{knockoffs} and it originates from the field of false discovery rate control in the setup of finding potential explanatory variables to an observed response variable. They are meant as a tool for estimating feature importance using conditional independence tests but have recently been applied as intervention variables for causal discovery \citep{Wasim_ICMLW}.

Let $u = \gamma \cdot f(\mathbf{Q}) + \eta$ be a predictive model for a vector of responses $\mathbf{u} \in \mathbb{R}^{n}$, an arbitrarily complex function $f$ of a known matrix $\mathbf{Q} \in \mathbb{R}^{n \times p}$ of potentially explanatory variables $(Q_{1}, \dots, Q_{p})$, an unknown vector of coefficients $\gamma$, and a Gaussian noise term $\eta \sim \mathcal{N}(0, \sigma^{2} \mathbf{I})$. One can then generate a Knockoff of any feature $Q_{j} \in \mathbf{Q}$ by first constructing a Gram matrix $\mathbf{\Sigma} = \mathbf{Q}^{T} \mathbf{Q}$, once all the features $Q_{j}$ are normalized, so that $\Sigma_{j,j}=|| Q_{j} ||_{2}^{2} = 1$ for all $j = 1, \dots, p$, and enforcing the following two conditions. First, the Knockoffs of $\mathbf{Q}$, denoted as $\tilde{\mathbf{Q}} = (\tilde{Q_{1}}, \dots, \tilde{Q_{p}})$, are constructed to have the same covariance structure as $\mathbf{Q}$ by enforcing $\tilde{\mathbf{Q}}^{T} \tilde{\mathbf{Q}} = \mathbf{\Sigma}$ and $\mathbf{Q}^{T} \tilde{\mathbf{Q}} = \mathbf{\Sigma} - \text{diag}(\mathbf{s})$, for a $p$-dimensional non-negative vector $\mathbf{s}$. Moreover, the inter-variable correlations between different original and Knockoff variables are enforced to be the same as those between the originals, that is, $Q_{j}^{T} \tilde{Q_{k}} = Q_{j}^{T}Q_{k}$ for all $j \neq k$. This condition is known as \textit{exchangeability}.

We rely on the work by \cite{deep_knockoffs} to construct approximate Knockoffs for arbitrary and unspecified distributions of the observational data. We chose the semi-definite programming (SDP) Knockoffs which are selected so that the original data and its Knockoff are as decorrelated as possible.

The Knockoffs are generated by using Gaussian mixture models as proposed by \cite{knockoff-gen}. First, the mixture assignment variable is sampled from the posterior distribution. The Knockoffs are then sampled from the conditional distribution given the original variables and the sampled mixture assignment such that the exchangeability condition is fulfilled \citep{knockoff-gen}.

\section{Data}\label{sec:data}

\subsection{Synthetic data}
\begin{figure*}[t]
	\centering
	\begin{minipage}{0.5\textwidth}
		\includegraphics[width=0.8\columnwidth]{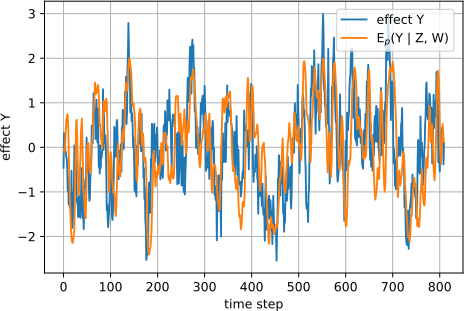}
		\label{fig:y0}
	\end{minipage}%
	\begin{minipage}{0.5\textwidth}
		\includegraphics[width=0.8\textwidth]{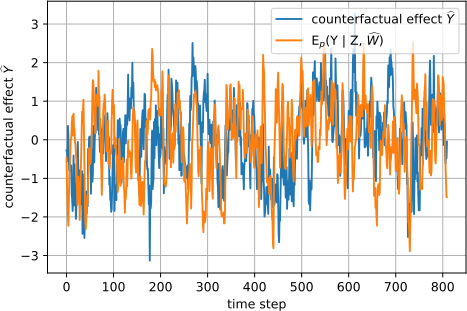}
		\label{fig:y1}
	\end{minipage}
	\caption{Left: Observational effect variable $Y$ from Eq. (\ref{eq:syn2_y}) (blue) and the conditional expectation of $Y$ given $Z$ and $W$ (orange). Right: Counterfactual $\hat{Y}$ (blue) of the effect variable $Y$, and the conditional expectation of $Y$ given $Z$ and $\widehat{W}$ (orange). Conditional expectations are estimated by SCEVAE with Knockoff intervention.}
	\label{fig:y0_y1}
\end{figure*}
Causal link estimation is impossible to validate unless we know the underlying system dynamics. For this reason we generate synthetic data. It contains a nonlinear causal link, all variables abide by the relations from DAG in Fig. \ref{fig:CEVAE_model} and form the structural causal model according to the following equations. By $z_{t}$, $x_{t}$, $w_{t}$, $\hat{w}_{t}$, and $y_{t}$ we denote the hidden confounder, its proxy, cause variable, and the effect variable at time step $t$, respectively.
\begin{align}
	z_{t} &= a \cdot z_{t-1} + {\epsilon_{1}}_{t} \label{eq:syn1_z}\\
	x_{t} &= b_{t} \cdot \tanh(z_{t-\tau}) + s \cdot {\epsilon_{2}}_{t} \label{eq:proxy}\\
	w_{t} &= c_{1} \cdot w_{t-1} + c_{2} \cdot z_{t} + {\epsilon_{3}}_{t} \label{eq:w}\\
	y_{t} &= d_{1} \cdot y_{t-1} + d_{2} \cdot z_{t} + g \cdot e^{h \cdot w_{t}} + {\epsilon_{4}}_{t}\label{eq:syn2_y}
\end{align}
In the case of Knockoff intervention $\tilde{w}_{t}$, the cause variable becomes $\hat{w}_{t} = do(w_{t} = \tilde{w}_{t})$, and in the Gaussian noise intervention case we have $\hat{w}_{t} \sim \mathcal{N}(0, \sigma^{2})$, for $\sigma^{2} \in \mathbb{R}$.
To asses how our method performs for estimating linear causal effects, we alter Eq. (\ref{eq:syn2_y}) as follows:
\begin{equation}\label{eq:syn1_y}
	y_{t} = d_{1} \cdot y_{t-1} + d_{2} \cdot z_{t} + g \cdot w_{t} + {\epsilon_{4}}_{t}
\end{equation}
In Eqs. (\ref{eq:syn1_z})-(\ref{eq:syn1_y}), $a, c_{i}, d_{i}, g, h, s, \sigma_{j}^{2} \in \mathbb{R}$, $i = 1, 2$, $\tau \in \mathbb{N}$, and noise terms $\epsilon_{j} = \mathcal{N}(0, \sigma_{j}^{2})$, for $j = 1, 2, 3, 4$, are mutually independent. By $e$ we denote the Euler's number. The time-varying proxy parameter $b_{t}$ (Fig. \ref{fig:time-varying_b}, Appendix \ref{appendix:add_figs}) is generated by setting $b_{t} = \frac{2}{N-2} t$ for $t=1, \dots, \frac{N}{2} - 1$ and symmetrically $b_{N-t-1} := b_{t-1}$ for $t=\frac{N}{2}, \dots, N$.

The counterfactual outcome $\hat{y}_{t}$ for data with nonlinear and linear causal link is obtained by substituting $w_{t}$ by $\hat{w}_{t}$ in Eq. (\ref{eq:syn2_y}) and Eq. (\ref{eq:syn1_y}), respectively.
The parameters used in our experiments are $a=0.88$, $s=0.5$, $\tau=100$, $c_{1}=0.6$, $c_{2}=0.2$, $d_{1}=0.4$, $d_{2}=0.8$, $g=0.5$, $h=0.5$, $\sigma^{2}=0.6$, $\sigma_{2}^{2}=1$, and $\sigma_{1}^{2}=0.5$, $\sigma_{2}^{2}=0.8$, $\sigma_{3}^{2}=0.7$, and $\sigma_{4}^{2}=0.5$. The hidden confounder and the cause variable are initialized with $z_{0}=0.1$ and $w_{0}=0.1$, whereas $y_{0}=0$.

\subsection{Cloud and aerosol observation data}

We demonstrate our method's performance in real-world scenarios by applying it to cloud and aerosol observations dataset provided by \cite{cloud_dataset}. It consists of the $1^{\circ} \times 1^{\circ}$ gridded version of the Moderate Resolution Imaging Spectroradiometer's measurements obtained twice per day at approximately $1$ km $\times$ $1$ km resolution from 2004 until 2019. Data stems from the Pacific Basin region off the coast of South America. As per \cite{uncertainty_clouds}, we use the daily average of all variables to downsample. As the outcome variable $Y$, we utilize cloud optical depth (COD) and as cause variable $W$ the aerosol optical depth (AOD). In line with \cite{uncertainty_clouds}, we use the following meteorological variables as proxies. Namely, sea surface temperature (SST), estimated inversion strength (EIS), vertical motion at 500mb (w500), relative humidity at 700mb, 850mb, and 900mb (RH700, RH850, RH900). All variables are normalized before we use them as input for training SCEVAE.

% Furthermore, we perform an additional experiment for comparison with \cite{uncertainty_clouds} in which we set $Y$ to COD, $W$ to AOD and use meteorological proxies RH700, RH850, RH900, EIS, w500, and SST. In this experiment, the data when precipitation values are greater than $0.05$ is discarded.

\section{Experiments}\label{sec:experiments}

Here we expound our experimental setup.
We generate $1000$-time-step-long multivariate time series and use the last $10\%$ of each variable as test data without shuffling in order to preserve inter-temporal dependencies. The last $10\%$ of training data is used for validation. We normalize the data by subtracting the mean of each variable and then dividing it by its standard deviation. We use batch size of $100$ and randomly select that many consecutive time steps as batches of the training data. The optimizer used is Adam \citep{Adam} with learning rate of $10^{-5}$, and a weight decay of $10^{-3}$. We model the hidden confounder as five-dimensional in the latent space in all but one experiment where we explicitly investigate the influence of its dimensionality to causal link estimation. The LSTMs of SCEVAE consist of two LSTM cells i.e. layers. Since the LSTMs that we use are not bidirectional, they can only look at the past time steps of a given time series. The hidden states and cell states of each LSTM have $32$ dimensions. We set the regularization parameter $\lambda$ from Eq. (\ref{eq:SCEVAE_var_lower_bound}) to $0.1$. For all experiments, we used GPUs of type GeForce RTX 2080 Ti.

\subsection{Results}\label{sec:results}

\subsubsection{Synthetic data experiments}
To demonstrate the efficacy of our proposed method by comparing its results to the ground truth causal link intensity, we apply it to synthetic data with either linear or nonlinear causal link between $W$ and $Y$.
Our method's reconstruction of the effect variable $Y$ and its counterfactual is $\hat{Y}$ is depicted in Fig. \ref{fig:y0_y1}. We qualitatively observe that the reconstructions (orange) are closely comparable to the corresponding observed values (blue). These variables are generated according to Eq. (\ref{eq:syn2_y}) using $W$ and $\widehat{W}$, respectively.

The quantitative results of SCEVAE's causal analysis are shown in Table \ref{fig:results}. The causality scores used are RMSE$_\text{ITE}$, denoting the RMSE between the ground truth and the predicted ITE as per Eq. (\ref{eq:ITE}), factual RMSE$_{Y}$, and counterfactual RMSE$_{\hat{Y}}$. The latter two scores measure the discrepancy between the observed $Y$ and the estimated $\mathbb{E}(Y | W, Z)$ at each time step $t$, and the discrepancy between $\hat{Y}$ and $\mathbb{E}(Y | \widehat{W}, Z)$ at each time step $t$, respectively. By "SCEVAE" in Table \ref{fig:results}, we denote our method with Gaussian noise intervention on $W$, whereas "SCEVAE-Knockoff" indicates that we used the Knockoff of $W$ as intervention $\widehat{W}$.
\begin{table*}[h]
	\centering
%	\vskip 0.15in
    \caption{Causal Effect Estimation Error Metrics for Linear and Nonlinear Causal Link of Synthetic Data for $g=0.5$. The results are averaged over five replications and shown with standard error. Lower is better.}
    \vskip 0.15in
	\begin{tabular}{c c c c c c}
		\toprule
		Causal link & parameter $b$ & Method & RMSE$_\text{ITE}$ & RMSE$_{Y}$ & RMSE$_{\hat{Y}}$\\
		\midrule
		\multirow{8}{*}{linear} & \multirow{4}{*}{time-varying} & SCEVAE & 0.52 $\pm$ 0.04 & 0.74 $\pm$ 0.06 & 0.99 $\pm$ 0.03\\
		& & SCEVAE-Knockoff & 0.34 $\pm$ 0.03 & 0.74 $\pm$ 0.01 & 0.63 $\pm$ 0.01 \\
        & & TSdeconf (without) & / & 0.75 $\pm$ 0.28 & 0.93 $\pm$ 0.41 \\
		& & TSdeconf (with) &/ & 0.75 $\pm$ 0.28 & 0.93 $\pm$ 0.41\\
		\cmidrule{3-6}
		& \multirow{4}{*}{0.95} & SCEVAE & 0.57 $\pm$ 0.06 & 0.72 $\pm$ 0.02 & 0.97 $\pm$ 0.03\\
		& & SCEVAE-Knockoff & 0.32 $\pm$ 0.01 & 0.76 $\pm$ 0.08 & 0.65 $\pm$ 0.07 \\
		& & TSdeconf (without) & / & 0.76 $\pm$ 0.29 & 0.93 $\pm$ 0.4 \\
		& & TSdeconf (with) & / & 0.75 $\pm$ 0.28 & 0.93 $\pm$ 0.41\\
        \midrule
		\multirow{8}{*}{nonlinear} & \multirow{4}{*}{time-varying} & SCEVAE & 0.56 $\pm$ 0.05 & 0.74 $\pm$ 0.03 & 0.99 $\pm$ 0.03\\
		& & SCEVAE-Knockoff & 0.44 $\pm$ 0.03 & 0.81 $\pm$ 0.03 & 0.7 $\pm$ 0.03 \\
		& & TSdeconf (without) & / & 0.85 $\pm$ 0.34 & 0.99 $\pm$ 0.47\\
		& & TSdeconf (with) & / & 0.84 $\pm$ 0.33 & 0.98 $\pm$ 0.46\\
		\cmidrule{3-6}
		& \multirow{4}{*}{0.95} & SCEVAE & 0.59 $\pm$ 0.1 & 0.79 $\pm$ 0.03 & 1.01 $\pm$ 0.01\\
		& & SCEVAE-Knockoff & 0.43 $\pm$ 0.03 & 0.8 $\pm$ 0.03 & 0.7 $\pm$ 0.03 \\
		& & TSdeconf (without) & / & 0.84 $\pm$ 0.33 & 0.98 $\pm$ 0.46\\
		& & TSdeconf (with) & / & 0.84 $\pm$ 0.33 & 0.98 $\pm$ 0.45\\
		\bottomrule
	\end{tabular}
	\label{fig:results}
\end{table*}
% In the linear case, as per Table \ref{fig:results}, the estimated causal link compared to the ground truth is ITE$=0.52$ when proxy parameter $b$ is time-varying and ITE$=0.57$ for a fixed value of parameter $b=0.95$. In the nonlinear case, for a time-varying parameter $b$ ITE$=0.56$, and for a fixed $b=0.95$ we obtain ITE$=0.59$. This means that our method can estimate the ground truth causal link intensity between $Y$ and $W$ similarly well in the nonlinear case as in the linear case but it performs better when the proxy parameter $b$ is time-varying.

According to the results from Table \ref{fig:results}, SCEVAE performs better in the linear than in the nonlinear causal link case, which illustrates the higher complexity of estimating nonlinear causal links.
% We also note that RMSE$_{\text{ITE}}$ is lower in the case of the time-varying parameter $b$ from Eq. (\ref{eq:proxy}), than when this parameter is fixed in the case of using the Gaussian noise as intervention on $W$.
We note that the methods works equally well for both fixed and time-varying values of the parameter $b$. Furthermore, in Table \ref{fig:results}, we can clearly see how using Knockoffs improves the causal link intensity estimation by remarkably lower RMSE$_{\text{ITE}}$ metric, as well as by inducing lower variance, making this a more reliable intervention choice. In addition, it is especially interesting to note that the counterfactual RMSE is much lower in comparison to other methods which do not use Knockoff interventions. This is due to the fact that Knockoff $\widehat{W}$ has the same distribution as $W$, and therefore produces a good reconstruction of the counterfactual outcome $\hat{Y}$ when we sample it from the learned estimate of $p(Y | W, Z)$.

We compare our results on synthetic data to those of Time Series Deconfounder (TSdeconf) \citep{Bica} either without taking hidden confounding into account or when the substitutes for the hidden confounder are generated and used as proxies, in Table \ref{fig:results} denoted by "TSdeconf (without)" and "TSdeconf (with)", respectively. We set the confounding parameter of TSdeconf to $\gamma = 0.8$, and the number of substitute, as well as the simulated hidden confounders to one. We note that TSdeconf only outputs RMSE between the ground truth and the predicted factual or counterfactual outcome, so these are the main comparison metrics. For obtaining the factual RMSE$_{T}$ via TSdeconf we input $X$ as covariates, $W$ as treatment, and $Y$ as outcome. To obtain the counterfactual RMSE$_{\hat{Y}}$, we also use $X$ as covariates, but $\widehat{W}$, and $\hat{Y}$ as treatment and outcome, respectively. Furthermore, since TSdeconf is not suitable for long time series, we generate $100$ $100$-time-step-long training samples for fairness. This is due to the fact that SCEVAE is trained on $100$ $100$-long epochs randomly chosen from our $1000$-time-step-long sequential variables.

We note that in the case of the linear causal link, TSdeconf performs almost as well as our method but with much higher variance, making our method considerably more stable. To obtain stable results using TSdeconf, one would need much larger amount of training data.
In the case of the nonlinear causal link, our method's superior performance becomes even clearer for both time-varying, and fixed proxy parameter $b=0.95$.
% \begin{figure}[t]
%     \centering
%     \includegraphics[width=0.4\columnwidth]{images/injective/ATE_train_test_vs_ATE_ground_truth.png}
%     \caption{Causal link intensity estimation. Ground truth ATE during training and test is shown by green and red dashed lines, respectively. Predicted ATE during training and testing is respectively shown in blue and orange.}
%     \label{fig:ATE_true_vs_pred}
% \end{figure}
% Counterfactual RMSE is higher than the factual RMSE since $\hat{Y}$ is sampled from the learned distribution of $Y$ using $\widehat{W}$ and was not learned individually.

% The results of causal link estimation using SCEVAE on synthetic data with nonlinear causal link are depicted in Fig. \ref{fig:ATE_true_vs_pred}. We see that our method converges to the ground truth ATE both during training and testing phases. The ground truth ATE values for both training and testing are shown in green and red dashed lines, respectively.
%We note that the uncertainty of the causal effect estimation during test increases over the epochs but on average test ATE is well-predicted. In contrast, the estimated ATE during training is much less uncertain.
% Furthermore, it is worth mentioning that the difference between the ground truth ATE during training and test stems from the test sample size since we choose the last $10\%$ of the original time series for testing where the mean of the entire variable might not always be accurately reflected.

The factual RMSE$_{T}$ of the outcome reconstruction during training and test can be found in Fig. \ref{fig:RMSE} (Appendix \ref{appendix:add_figs}). In Fig. \ref{fig:z_coeff_exp} (Appendix \ref{appendix:add_figs}) we show how dimensionality of $Z$ in the latent space and the confounding coefficient $d_{2}$ influence the factual and counterfactual outcome's reconstruction. We tested latent dimensions $D_{Z} \in \{1, 5, 10, 20\}$, and coefficients $d_{2} \in \{0.8, 1, 1.2, 1.6\}$. We observe that RMSE values in both factual and counterfactual cases are lowest when $d_{2}=0.8$ and have an upward trend as the rate of hidden confounding $d_{2}$ increases regardless of $D_{Z}$. As the dimensionality of $Z$ increases, the uncertainty of the prediction becomes higher.

\subsubsection{Cloud and aerosol data experiments}

In the experiments on real aerosol-cloud-climate observations, where we use COD as the effect and AOD as the cause variable, we demonstrate the importance of choosing suitable proxies. Since now we do not have the ground truth causal link intensity values, we cannot use RMSE$_{\text{ITE}}$. Instead, we use ATE metric as per Eq. (\ref{eq:ATE}) of the predicted factual and counterfactual outcome variables $Y$ and $\hat{Y}$, respectively. Moreover, we note that intervention on real data is often not feasible in practice, but that by our method's way of intervention we are able to counterfactually analyse the desired causal link.
\begin{table*}[h]
	\centering
    \caption{Causal Effect Estimation Error Metrics for Cloud-Aerosol Dataset when using COD as Outcome Variable and AOD as Intervention Variable. The results are averaged over five replications and shown with standard error.}
    \vskip 0.15in
	\begin{tabular}{c c c c c}
		\toprule
		Method & Proxy & ATE train & ATE test & RMSE$_{Y}$\\
		\midrule
		\multirow{2}{*}{SCEVAE} & meteorological & 0.06 $\pm$ 0.05 & 0.09 $\pm$ 0.08 & 0.72 $\pm$ 0.001\\
		& uniform & 0.21 $\pm$ 0.18 & 0.24 $\pm$ 0.14 & 0.79 $\pm$ 0.14 \\
		\cmidrule{2-5}
% 		\multirow{2}{*}{SCEVAE-Knockoff} & meteorological & 0.21 $\pm$ 0.13 & 0.14 $\pm$ 0.09 & 0.95 $\pm$ 0.13 \\
		\multirow{2}{*}{SCEVAE-Knockoff} & meteorological & 0.03 $\pm$ 0.02 & 0.04 $\pm$ 0.02 & 0.78 $\pm$ 0.02 \\
% 		\multirow{2}{*}{SCEVAE-Knockoff} & RH700 & 0.05 $\pm$ 0.02 & 0.02 $\pm$ 0.01 & 0.9 $\pm$ 0.004 \\
		& uniform & 0.29 $\pm$ 0.08 & 0.37 $\pm$ 0.06 & 0.83 $\pm$ 0.03 \\
		\bottomrule
	\end{tabular}
	\label{fig:cloud_results}
\end{table*}

In case of \textit{meteorological} proxy, we set it to be multivariate including SST, EIS, w500, RH700, RH850, and RH900 as per \cite{uncertainty_clouds}. Whereas when the proxy is indicated as \textit{uniform}, we set it univariately to $\mathcal{U}(0,1)$. The results of varying the choice of proxy $X$ are shown in Table \ref{fig:cloud_results} for SCEVAE with standard normal, as well as with Knockoff intervention.

We note that using meteorological proxies yields lower variance of both training and test ATE in contrast to the case where we used uniform noise as the proxy. Moreover, SCEVAE with Knockoff intervention yields lower RMSE$_{Y}$ with the use of meteorological proxies. This implies the importance of the appropriate proxy choice.

% Furthermore, we note that the true causal link between COD and AOD measured by ATE in the case of meteorological proxies is weaker when we use SCEVAE with Knockoff, than when it is used with standard normal intervention.
Furthermore, when using meteorological proxies the causal link intensity between COD and AOD is lower than in the case we set $X$ to uniform noise during both training and test. Similar findings, i.e. that aerosol has limited impact on cloud depth, were reported by \cite{uncertainty_clouds} for a comparable rate of confounding. This further strengthens our claim that choosing suitable proxies is crucial for correctly estimating causal link intensity under hidden confounding as it may contribute to identifiability.

\section{Conclusion}\label{sec:conclusion}

Causal link intensity estimation is a challenging task, especially in the presence of latent confounders. In this paper we introduced SCEVAE, a novel deep learning method for time series causality analysis under hidden confounding to tackle this problem in sequential data with Knockoff interventions. It is inspired by the CEVAE framework, but applicable to complex non-stationary time series through the use of LSTMs and fundamental architectural novelty. Our method allows for single-variable causality analysis instead of using many independent samples for training. We achieved better and more stable results than the time series deconfounding benchmark TSdeconf. Moreover, we showed that estimating the confounded causal link intensity can be done more accurately with Knockoff rather than using standard normal interventions. This was attained on synthetic data with both linear and nonlinear causal links. In addition, we observed that using Knockoff interventions reduces the counterfactual RMSE$_{\hat{Y}}$ in comparison to SCEVAE without Knockoff intervention since the distribution of the Knockoff agrees with the learned distribution of the cause variable. Moreover, through the experiments on real cloud-aerosol observational data, we indicated our method's potential applicability to real-world problems and illustrated how the use of meaningful proxies contributes to its identifiability.

% \section*{References}
\section*{Acknowledgements}

Funding for this study was provided by the German Research Foundation (DFG) individual research grant SH 1682/1-1 and the European Research Council (ERC) Synergy Grant “Understanding and Modelling the Earth System with Machine Learning (USMILE)” under the Horizon 2020 research and innovation programme (Grant agreement No. $855187$). 

\bibliographystyle{authordate1}
\bibliography{references}

\newpage

\appendix

\section{Appendix}\label{appendix:add_figs}

\begin{figure}[h]
    \begin{minipage}{.5\textwidth}
		\centering
		\includegraphics[width=0.7\columnwidth]{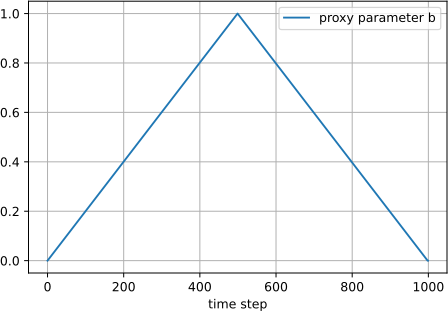}
		\caption{Time-varying proxy parameter $b_{t}$ \\
		over $1000$ time steps.}
    \label{fig:time-varying_b}
	\end{minipage}%
	\begin{minipage}{.5\textwidth}
		\centering
		\includegraphics[width=0.7\columnwidth]{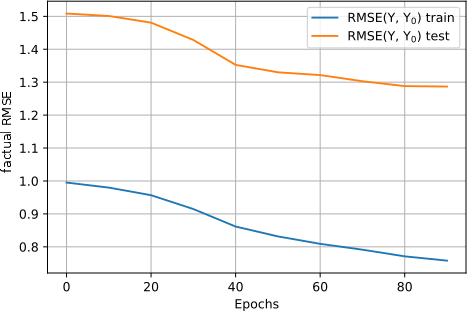}
		\caption{RMSE during training (blue) and test (orange) for the synthetic data with nonlinear causal link and time-varying proxy parameter $b$.}
	\label{fig:RMSE}
	\end{minipage}
\end{figure}

% \begin{figure}
%     \centering
%     \includegraphics[width=0.7\columnwidth]{images/y0_uniform_noise_proxy.png}
%     \caption{Outcome prediction when proxy is set to uniform $\mathcal{U}(0,1)$ noise instead of using meteorological proxies for cloud-aerosol data.}
%     \label{fig:noise_proxy_clouds}
% \end{figure}

% \begin{figure*}[h]
% 	\begin{minipage}{.5\textwidth}
% 		\centering
% 		\includegraphics[width=0.7\columnwidth]{images/injective/w_x.png}
% 	\end{minipage}%
% 	\begin{minipage}{.5\textwidth}
% 		\centering
% 		\includegraphics[width=0.7\columnwidth]{images/injective/w_z.png}
% 	\end{minipage}
% 	\caption{Left: Reconstruction of the cause variable $W$ by the proxy variable $X$. Right: Reconstruction of $W$ by approximated $Z$. For both figures we applied our method to synthetic data with a nonlinear causal link.}
% 	\label{fig:reconstruction_w}
% \end{figure*}

\begin{figure*}[h]
	\begin{minipage}{.5\textwidth}
		\centering
		\includegraphics[width=0.7\columnwidth]{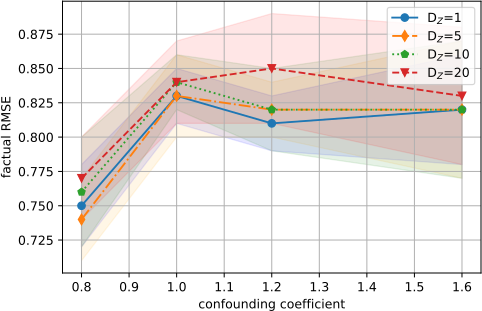}
	\end{minipage}%
	\begin{minipage}{.5\textwidth}
		\centering
		\includegraphics[width=0.7\columnwidth]{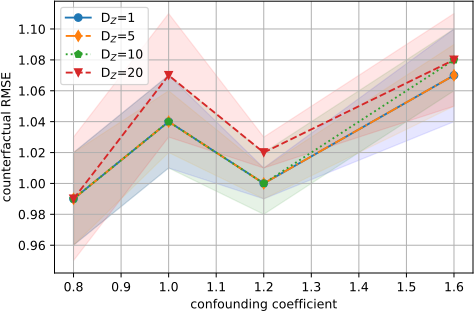}
	\end{minipage}
	\caption{Influence of latent dimension and coefficient of the latent confounder to the outcome forecast with standard normal intervention. Factual RMSE is shown on the left and counterfactual RMSE on the right with standard deviation after five replications.}
	\label{fig:z_coeff_exp}
\end{figure*}

\end{document}